\documentclass[10pt,amssymb,twocolumn,notitlepage]{article}

\usepackage{epigraph}
\usepackage[affil-it]{authblk}
\usepackage{dcolumn}
\usepackage[stable]{footmisc}
\usepackage{blkarray}
\usepackage{bm}
\usepackage[mathscr]{euscript}
\usepackage[font=scriptsize]{caption}
\usepackage{subcaption,graphicx}
\usepackage[table]{xcolor}
\usepackage[margin=1.7cm]{geometry}
\usepackage{tcolorbox}
\usepackage{hyperref}
\usepackage{gensymb}
\usepackage[font=footnotesize]{caption}
\usepackage{hyperref}
\hypersetup{
    colorlinks=true,
    linkcolor=blue,
    filecolor=magenta,      
    urlcolor=blue,
}
\usepackage{steinmetz}
\usepackage{amssymb}
\usepackage{makecell}
\usepackage{algorithm,lipsum,xcolor,caption}
\usepackage{enumitem}
\usepackage{amsmath,esint}
\usepackage{fancyvrb}
\usepackage{xcolor}
\usepackage{arydshln}
\usepackage{mathtools}

\definecolor{newblue}{rgb}{0.0, 0.28, 0.67}
\definecolor{newgreen}{rgb}{0.13, 0.55, 0.13}
\definecolor{newred}{rgb}{0.87, 0.72, 0.53}

\definecolor{newblue}{rgb}{0.0, 0.28, 0.67}
\definecolor{newgreen}{rgb}{0.13, 0.55, 0.13}
\definecolor{newred}{rgb}{0.87, 0.72, 0.53}
\usepackage{stackengine}

\title{Comparing Cross Correlation-Based Similarities}
\author{Luciano da Fontoura Costa \\ \emph{luciano@ifsc.usp.br}}
\affil{S\~ao Carlos Institute of Physics -- DFCM/USP} 
\date{21st Oct 2021}

\begin{document}

\twocolumn[
\begin{@twocolumnfalse}
    \maketitle
    \begin{abstract}
The real-valued Jaccard and coincidence indices, in addition to their
conceptual and computational simplicity, have been verified to be able to provide promising results in  
tasks such as template matching,  tending to yield peaks that are sharper and narrower than those typically obtained by standard cross-correlation, while also attenuating substantially secondary matchings.  In this
work, the multiset-based correlations based on the real-valued multiset Jaccard and coincidence indices are compared from the perspective of template matching, with encouraging results which have implications for pattern recognition, deep learning, and scientific modeling in general.   The multiset-based correlation methods, and especially the coincidence index, presented remarkable performance characterized by sharper and narrower peaks while  secondary peaks were attenuated, which was maintained even in presence of intense levels of noise.  In particular, the two methods derived from the coincidence index led to particularly interesting results.  The cross correlation, however, presented the best robustness to symmetric additive noise, which suggested a new  combination of the considered approaches.  After a preliminary investigation of the relative performance of the multiset approaches, as well as the classic cross-correlation, a systematic comparison framework is proposed and applied for the study of the aforementioned methods.  Several results are reported, including the confirmation, at least for the considered type of data, of the coincidence correlation as providing enhanced performance regarding detection of narrow, sharp peaks while secondary matches are duly attenuated.  The combined method also resulted promising for dealing with signals in presence of intense additive noise.
     \end{abstract}
\end{@twocolumnfalse} \bigskip
]

\setlength{\epigraphwidth}{.49\textwidth}
\epigraph{`\emph{Deep inside the mirror, a whole universe of similarities.}'}
{\emph{LdaFC}}

\section{Introduction}

Many concepts and methods in science, and in particular in Physics, rely on 
operations capable of expressing the similarity between two mathematical
structures or data.  Of particular interest and importance has been the inner
product between two vectors or functions, which underlies a vast quantity
of concepts and results in Physics, especially when incorporated into the
convolution and correlation operations.

The aforementioned concepts and operations play a particulary important
role in optics, mathematical physics, semiconductors,
complex systems, quantum mechanics and computing,  magnetic resonance, astrophysics,  
and  neuronal networks, to name but a few possibilities.  

The Jaccard index (e.g.~\cite{Jaccard, Jac:wiki, CostaJaccard}) has been 
extensively used in many areas as an interesting and
effective means for quantifying the similarity between any two sets, mainly with respect to binary
or categorical data.   A multiset Jaccard index, in which
multisets (e.g.~\cite{Hein,Knuth,Blizard,Blizard2,Thangavelu,Singh}) are taken instead of sets, 
has also been used in
similar applications, with the ability to take into account the multiplicities of
multiset elements.   More recently~\cite{CostaJaccard,CostaMset,CostaSimilarity}, 
several further generalizations  of the Jaccard index have been proposed.  
 By taking into account real-valued data, these
developments also allowed the multiset concepts and operations to be extended to
real functions and signals, which have been called multifunctions.
   
As with its real function counterpart, the multifunction correlation (and convolution) can be employed for
typical signal processing tasks, including but not being limited to filtering, template matching 
(e.g.~\cite{Schalkoff,shapebook}),
and control theory. Indeed, preliminary results~\cite{CostaJaccard,CostaMset} 
have been obtained that are particularly promising and encouraging.  More specifically, when
used for template matching, the obtained peaks corresponding to the maximum similarity between the
object and template functions are not only substantially sharper and narrower, but also the secondary
matching peaks result much more attenuated~\cite{CostaMset}. 

The combination of these two features suggests that the  multifunction convolution has substantial 
potential for applications in pattern recognition, neuronal networks and deep learning, as well as in
many other related areas involving estimations of similarity and/or convolutions.  

For all the above reasons, it becomes of particular important to compare, in a quantitative and
objective manner, the performance of the several mentioned correlation methods based on
the multiset-based similarity indices, including the real-valued Jaccard and coincidence indices~\cite{CostaJaccard}.  The classic cross-correlation operation should  also be taken into
account as a reference method, given its extensive application in the most diverse areas.
This constitutes the main purpose of the present work.  

We start by reviewing the adopted indices, and then proceed to a preliminary comparison 
related to template matching between an object and a template function in presence of noise, in
which it is identified that the addition-based Jaccard and respectively associated addition-based
coincidence present performance similar to the real-valued Jaccard and coincidence indices.
Though the classic cross-correlation led to not particularly noticeable matching results, it has been 
verified to be more resilient to elevated levels of noise than the other
considered cross-correlation methods.
Given that the classic cross-correlation allowed good robustness to high levels of noise
relatively to the similarity index-based methods, we also propose a combine method
in which the former approach is applied prior to the latter indices.

Guided by the aforementioned preliminary comparison, attention is subsequently focused 
on the real-valued Jaccard and coincidence
indices, as well as on the classic cross-correlation product.  A systematic framework for
comparing similarity cross-correlations is then proposed that can be understood as an
ancillary contribution of the present work.  Several interesting results are obtained, including
the identification of the coincidence method as the most effective among the considered
approaches, at least for the type of
signals considered here.     In addition, the combined method confirmed its
potential for enhanced performance when the signals contain high levels of noise, especially
for smooth template functions.

We start by presenting the considered cross-correlation methods, and then report on the
preliminary comparison.  The formal, systematic framework for comparing cross-correlation
methods is then presented and applied for respective characterization of the considered
cross-correlation approaches.

\section{The Considered Similarity Indices}

The \emph{basic Jaccard index} between two sets with non-negative multiplicity can be defined as:
\begin{equation}  \label{eq:Jac}
   \mathcal{J}(A,B) =   \frac{\left| A \cap B \right|}  {\left| A \cup B \right|}
\end{equation}

where $A$ and $B$ are any two sets to be compared.  It can be verified that
$0 \leq  \mathcal{J}(A,B) \leq 1$.

The \emph{homogeneity} or \emph{interiority index} (also known as overlap index e.g.~\cite{Kavitha})  can be expressed as:
\begin{equation} \label{eq:Int}
   \mathcal{I}(A,B) =   \frac{\left| A \cap B \right|}  {\min \left\{ \left| A \right|, \left| B \right| \right\} }
\end{equation}

with $0 \leq \mathcal{I}(A,B) \leq 1$ and $0 \leq  \mathcal{J}(A,B) \leq \mathcal{H}(A,B)  \leq 1$.

The combination of the basic Jaccard and the interiority indices yields the 
\emph{coincidence index}, proposed in ~\cite{CostaJaccard} as:
\begin{equation}
    \mathcal{C}(A,B) = \mathcal{J}(A,B) \; \mathcal{I}(A,B) 
\end{equation}

The Jaccard index extended to multisets with non-negative multiplicities can be expressed as:
\begin{equation}  \label{eq:Jac_multi}
   \mathcal{J}_M(A,B) =   \frac{\sum_{i=1}^N\min{(a_i,b_i)}}  {\sum_{i=1}^N\max{(a_i,b_i)}} 
 \end{equation}

with $0 \leq \mathcal{J}_M(A,B) \leq$ and where $a_i$ and $b_i$ are the multiplicities of the
involved elements.  

The extension of the Jaccard index to real, possibly negative-valued multiplicities, involves
the following developments~\cite{CostaJaccard,CostaMset}.  
In particular, the possibility to have positive and/or negative multiplicity values 
now requires the application of
a binary operator analogous to the inner product in function spaces, in which the multiplicities are
properly mirrored among the four quadrants depending on their signs.  Previous related
developments include~\cite{mirkin,Akbas1,Akbas2}.

In particular, we adopt the sign scheme in~\cite{mirkin} as we define the following functional:
on the two functions $f(x)$ and $g(x)$:
\begin{equation}
   s_+ = \int_S |s_f + s_g| / 2 \min\left\{ s_f f, s_g g \right\} dx
\end{equation}

where $S$ is the combined support of the two functions, and $s_x = sign(x)$, $s_y = sign(y)$.

It is also possible to consider:
\begin{equation}
   s_- = \int_S |s_f - s_g| / 2 \min\left\{ s_f f, s_g g \right\} dx
\end{equation}

and then combine these two functionals~\cite{CostaSimilarity} as:
\begin{equation}
   s_{\pm} = \left[ \alpha \right] s_+ - \left[ 1 - \alpha \right] s_-
\end{equation}

where the parameter $\alpha$, with $0 \leq \alpha \leq 1$, 
controls the weight of the contributions of portions of the
signals that have the same or opposite signals.

When $\alpha=0.5$ we get:
\begin{equation}
   f \sqcap g = \int_{S}  s_f s_g \min(s_f f(x), s_g g(x)) dx 
\end{equation}

which is the sign scheme originally described in~\cite{Akbas1,Akbas2}.  
This functional has also been verified to be directly related to the intersection between 
real-valued multisets~\cite{CostaGenMops}.

We shall also adopt the following operator, which is related to the absolute union of real-valued
multisets~\cite{CostaGenMops}:
\begin{equation}
    f(x) \tilde{\sqcup} g(x)  = \int_{S}   \max(s_f f(x), s_g g(x)) dx 
\end{equation}

We can now define the following functional:
\begin{equation}
   \mathcal{J}_R(f,g) = \frac{  \int_{S}  s_f s_g \min(s_f f(x), s_g g(x)) dx }
   { \int_{S}   \max(s_f f(x), s_g g(x)) dx}
\end{equation}

which corresponds to the real-valued Jaccard similarity index~\cite{CostaJaccard,CostaSimilarity}.

Now, the  a possible normalized Jaccard correlation can be more compactly expressed as:
\begin{equation}  \label{eq:negJac}
  \mathcal{J}_R(f,g) [y] =  \frac{ \int_{S}   f(x) \sqcap g(x-y) dx } { \int_{S} f(x) \tilde{\sqcup} g(x-y) dx}
\end{equation}

The interiority index for real-valued multisets can be expressed as:
\begin{equation} \label{eq:Int}
   \mathcal{I}(f(x),g(x)) =  \frac{ \min\left\{ s_f f(x), s_g g(x) \right\} } {\min\left\{ A_f, A_g \right\}}
\end{equation}

where:
\begin{eqnarray}
    A_f =  \int_{S}  |f(x)| dx   \\
    A_g =  \int_{S}  |g(x)| dx 
\end{eqnarray}

The respective product between the real-valued Jaccard and interiority index yields the
\emph{coincidence index}~\cite{CostaJaccard,CostaMset}, and
 respective correlation and convolution.

The real-valued Jaccard index can be adapted for taking into account the
\emph{sum} of the two sets $A$ and $B$,  instead of their respective union, which
leads to the \emph{addition-based real-valued Jaccard index}:
\begin{equation}  \label{eq:addJac}
  \mathcal{J}_A(f(x),g(x)) =  \frac{2 \int_{S} \left(f \sqcap g \right) dx}  {\int_{S} \left( f(x)+g(x)  \right) dx}
\end{equation}

with $0 \leq \mathcal{J}_S(f(x),g(x)) \leq 1$.

The addition-based real-valued Jaccard index can also be
combined with the interiority index, leading to the respective \emph{addition-based coincidence
index}.

Each of these indices lead to respective multifunction convolutions and correlations, 
which involve sliding one function with respect to the other while calculating the respective index,
followed by the respective integration.  

Several other indices can be derived from those
described above by choosing other functionals for numerator and denominator of the
Jaccard index, as well as by taking into account other product combinations of those
indices.

\section{Preliminary Comparison} \label{sec:prelim}

\begin{figure*}[h!]  
\begin{center}
   \includegraphics[width=0.9\linewidth]{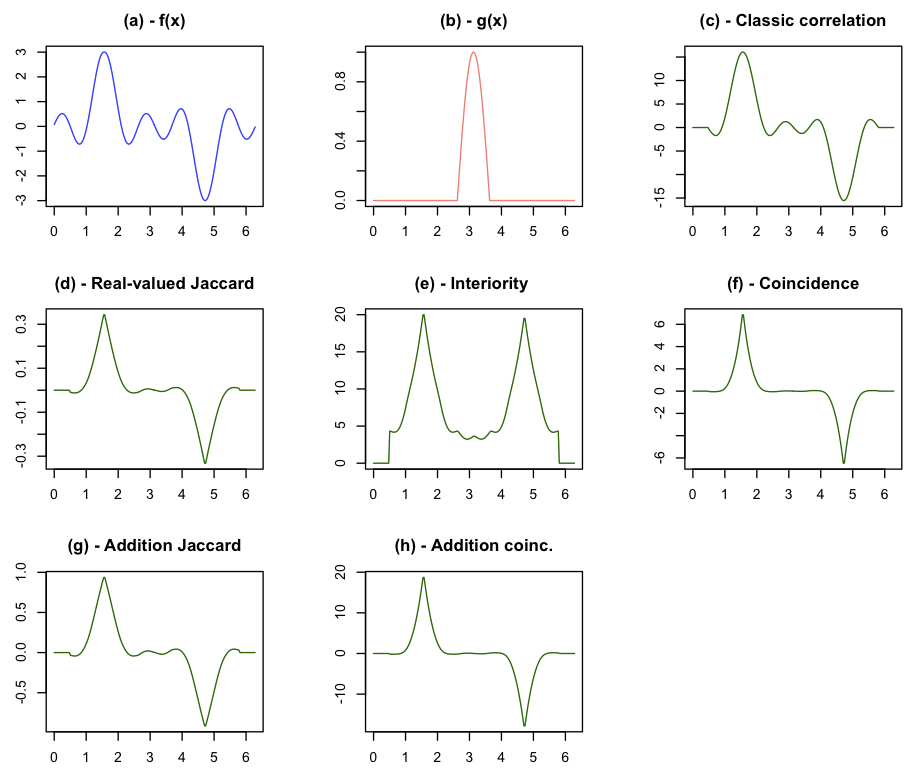}  
    \caption{Comparison of the several  multifunction convolution methods with respect to
    two noiseless functions (a) and  (b). }
    \label{fig:n_0}
    \end{center}
\end{figure*}

In order to have a first indication about the relative performance of all the considered methods 
in presence of increasing noise levels, we performed 
a respective preliminary comparison regarding the object and template functions
shown in Figure~\ref{fig:n_0}(a) and (b), respectively.   Also shown in that
figure are the results of results of applying the several types of correlations/convolutions
considered in the present work in the case of null noise.  

\vspace{0.5cm}

As expected, the multiset-based correlations yielded substantially enhanced
results regarding the identification of the peaks corresponding to the matches, leading
to sharper and narrower peaks and attenuation of secondary matches, while the standard 
cross correlation resulted with peaks that are even wider than in the original object.  

Of particular interest is to observe the results of the interiority-based convolution, 
yielding two sharp positive peaks that are,
however, less narrow than those obtained for the other multifunction convolutions.
This interesting property contributed to the verified enhanced performance of the
two coincidence index-based multifunction convolution methods, which yielded the
sharpest and narrowest matching peaks while almost eliminating the secondary
matches.

Subsequent results considered the incorporation of progressive noise levels
into the object function.  More specifically, we added noised points at each of the
$x$ values drawn from the symmetric, uniform density:
\begin{equation}
  n(x) = L (u(x)-0.5)
\end{equation}

where $u()$ is the uniform random distribution in the interval
$[0,1]$ and $L$ is the \emph{noise level}.

The best results, in the sense of enhancing the main peak
while attenuating the secondary matches, resulted from the application of the coincidence
and addition-based coincidence correlations.

It follows from the analysis of these results that all considered methods are robust
to substantially high levels of noise.   Actually, the multiset based approaches results 
combine a high-pass action in sharpening and enhancing the peaks with a
low-pass effect in substantially reducing the high-frequency noise.

However, two effects are of particular importance.
First, we have that the added noised implied in slight loss of sharpness in the case
of all the multiset methods.  At the same time, the classic cross correlation accounted
for the best robustness to noise.  These complementary features motivated the
combination of these two types of methods as described in the following section.

\section{Combining Cross Correlation and Multifunction Correlations} \label{sec:comb}

We have seen that, while the multiset-based cross-correlation methods allow enhanced
performance compared the classic cross-correlation, the latter is more robust to the
highest considered noise levels.  These complementary features of these two families
of methods motivate us to propose a combined method that would take advantage
of both interesting features, to be applied in elevated levels of noise.

This can be immediately achieved by applying the multiset methods \emph{after}
the two signals have been cross-correlated, with the result being taken as the next
object function.  However, the noise tolerance will depend on the smoothness of the
template function.  It is also possible to apply low-pass filtering on the noise functions, but this
may not be necessary in case the template function is itself smooth as in the case of the
current examples.

\begin{figure*}[h!]  
\begin{center}
   \includegraphics[width=0.9\linewidth]{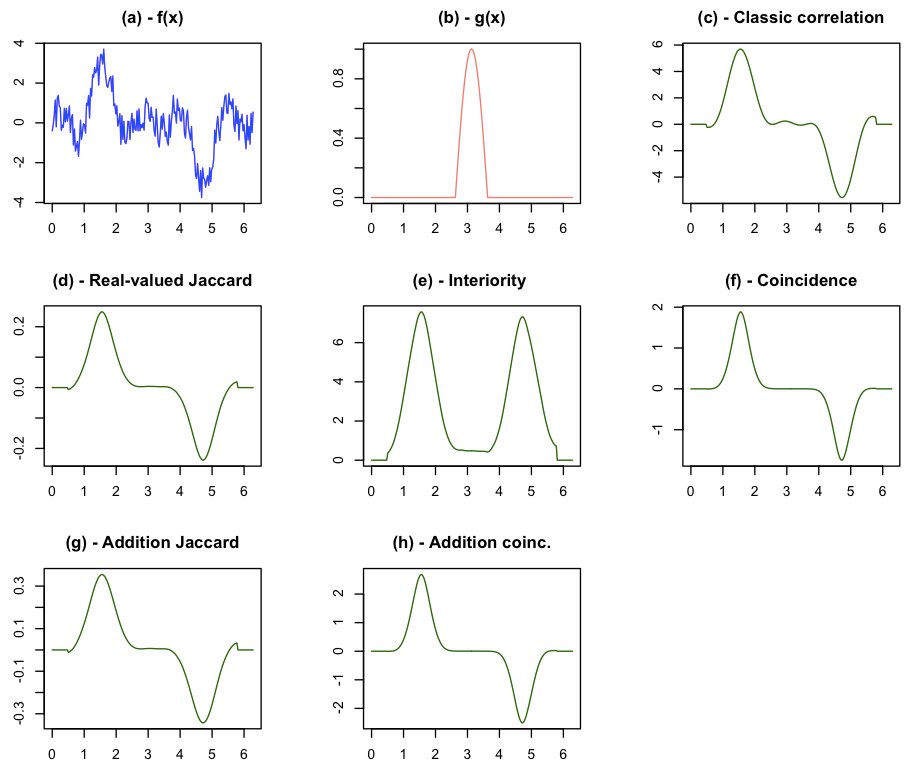}  
    \caption{Results of the combination of cross-correlation with multifunction convolutions.
    The cross correlation was first obtained between the object
    signal with level noise $L=2$ and respective template function, 
    then followed by the several considered methods.  It can be readily observed that
    this respective methodology allowed the combination of the good characteristics of
    the classic cross correlation (higher robustness to noise) with the enhanced
    peaks provided by the multifunction convolutions considered in the present work. }
    \label{fig:comb}
    \end{center}
\end{figure*}

Figure~\ref{fig:comb} illustrates the results obtained by application of the above proposed
methodology on the functions in Figure~\ref{fig:n_0}, a situation which corresponds to the
highest levels of noise considered in this work.  The obtained results corroborate the 
potential of the approach.

\vspace{0.5cm}

\section{The Comparison Framework}

In order to compare the performance of the considered correlation methods in an objective
and systematic manner, it
is first necessary to develop a formal and comprehensive respective framework taking into
account all the parameters of the object and template function, as well as
defining several suitable quantifications of the several performance aspects of interest.

The preliminary performance investigation reported in the previous section  
contributed to obtaining a more effective and objective systematic approach.  For instance,
since the real-valued addition-based Jaccard and respective coincidence indices were
fount to yield performance very close to the real-valued Jaccard and coincidence indices
regarding the addressed operation of template matching, only the two latter indices are
taken into account in a more systematic manner.  Also, given that the interiority index
is focused on providing a partial indication of the similarity between the template and
object functions, related mainly to the relative interiority between these two functions, the
interiority index will also not be considered further in our subsequent performance
evaluation.  

However, despite its almost hopeless performance of the classic cross-correlation
approach, it will be considered as a reference because of its extensive application in the
most diverse areas.

\begin{figure}[h!]  
\begin{center}
   \includegraphics[width=0.9\linewidth]{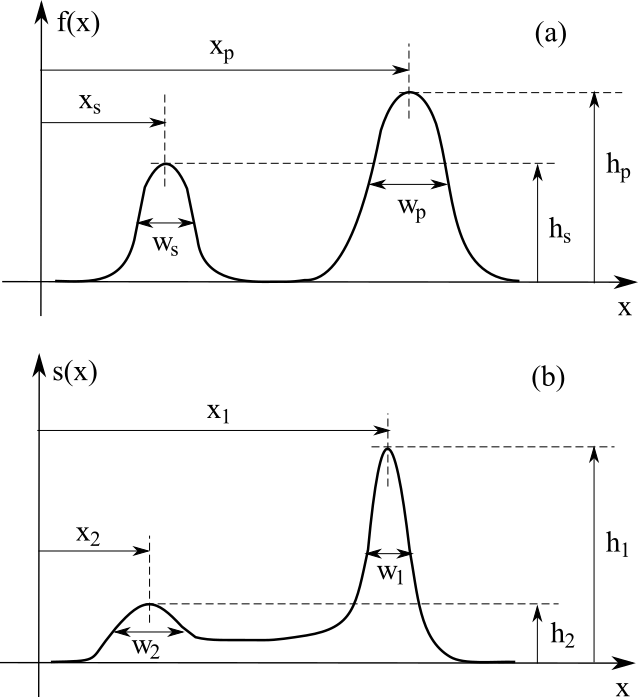}  
    \caption{The basic framework for comparison of correlation methods.  The
    object function has two peaks, one principal ($p$) and another secondary $s$).
    The result of the correlation yields two peaks, one corresponding to the principal
    original peak ($1$), and the other to the secondary original peak ($2$).}
    \label{fig:framework}
    \end{center}
\end{figure}

Figure~\ref{fig:framework} depicts the basic framework proposed in this work
for systematic comparison of the performance of the application of the
considered correlation methods to the task of template matching.

\vspace{0.5cm}

The henceforth adopted object function in (a), corresponding to the addition of two
gaussians, incorporates all the aspects of interest in our comparison approach.
Here, we have a \emph{principal} peak with height $h_p$ and width $w_p$ together
with a \emph{secondary} peak with respective $h_s$ and $w_s$.  Each of these
two gaussians are placed at respective abscissae values $x_p$ and $x_s$.

The expected cross-correlation result is to emphasize as much as possible the dominant 
peak while the secondary match is attenuated.  More precisely, the ideal result would correspond
to a single Dirac's delta peak at position $x_p$.  However, the typical expected result will be
as indicated in Figure~\ref{fig:framework}(b).  The main resulting peak will be positioned
at $x_1$, having height $h_1$ and width $w_1$.  A secondary peak is likely to be
obtained at respective position $x_2$, with height $h_2$ and width $w_2$.
In addition, an overlap region extending from $x_2$ to $x_1$ is also likely to be
obtained as result of the cross-correlation methods.  The shapes of the detected
peaks will typically be different from those of the respective original counterparts.

The width of the four peaks is determined by measuring the extension of the
respective slice at $75\%$ of the respective height.  We have not adopted the
standard deviation (other for initial specification of the peaks) because, though it would 
be viable regarding the object
function in (a), it is impossible to be properly characterized from the resulting
matching profile $s(x)$ in (b), as this would require some criterion and
respective methodology for separating the two typically overlapping peaks.

We can take the following interesting additional parameters regarding the object function
that can be used while comparing and discussing the correlation methods:
\begin{align}
      &\Delta x = x_p - x_s   \\
      &r_h = \frac{h_p} {h_s}  \\
      &r_w =  \frac{w_s}{w_p}
\end{align}

The first parameter, $\Delta x$, specifies  the separation between the two peaks.
The ratio $r_h$ indicates how secondary the smaller peak is, while the 
ratio $r_w$ expresses the relationship between the two widths.  These parameters
correspond to those that are most likely to influence the performance of the
correlation approaches.   Good results will be favored by relatively large values of
$\Delta x$ and $r_h$.  The influence of the ratio $r_w$ is not so straightforward.

Now that we have identified all relevant parameters, we are in position to 
objectively define figures of merit regarding the respective
performance.   Consider the prototypical matching function depicted in
Figure~\ref{fig:framework}(b).  The adopted performance indicators are as follows:
\begin{equation} 
\begin{aligned}
      &r_{ x_p} = \frac{x_p - x_1}{x_p}   &\emph{desirable low}   \nonumber \\
      &r_{ x_s} = \frac{x_s - x_2}{x_s}   &\emph{desirable low} \\
      &r_h =  \frac{h_1/h_2}  {h_p/h_s}     &\emph{desirable high} \\
      &r_{w,p} =  \frac{w_1}{h_1}              &\emph{desirable low} \\
      &r_{w,s} =  \frac{w_2}{h_2}              &\emph{desirable high} \\
      &\alpha = \int_{x_2}^{x_1}  s(x) dx    &\emph{desirable low}
\end{aligned}
\end{equation}

The ratio $r_{x_p}$ quantifies how much the main resulting peak displaces itself
from the original principal peak.  As such, this corresponds to a relative error and
should be as small as possible in order to ensure effective template detection.
The ratio $r_{x_s}$ plays an analogue role respectively to the secondary peaks, 
and therefore also should be as small as possible.

The relationship between the relative heights of the main and secondary original
and detected peaks is quantified by the relative index $r_h$.  The higher this value, 
the most the main peak will differentiate in height from the secondary peak.
This immediately implies that the secondary peak will result relatively more attenuated.
As such, this index should be as high as possible.

The ratio $r_{w,p}$ quantifies how narrow the obtained detected peak is and,
as such, should be as small as possible.    The index $r_{w,s}$ plays a similar
role regarding the secondary peak, and as such should be as high as possible.

The index $\alpha$ corresponds to the integral of the obtained matching function
$s(t)$, therefore quantifying how intense the resulting overlap is.  Ideally, we should
have $\alpha=0$, so that the lower this value, the better the performance will be.

\section{Performance in Presence of Additive Noise}

The comprehensive systematic framework developed in the previous section is now
applied as a means for comparing the similarity-based correlation approaches.
More specifically, we will incorporated progressive levels of additive noise to the
object function as in Figure~\ref{fig:framework}, with the following parameters:
\begin{align}
  &\sigma_p = 0.3 \nonumber \\
  &\sigma_s = 0.15 \nonumber \\
  &h_p = 2 \nonumber \\
  &h_s = 1 \nonumber \\
  &x_p = 4.5 \nonumber \\
  &x_s = 1.8 \nonumber
\end{align}

The 21 noise levels are as follows:
\begin{equation}
  ns\left[v\right] = \frac{v}{20} \left[ u(x) - 0.5\right], \emph{ with } v = 0, 1, \ldots, 20
\end{equation}

While the first level correspond to noiseless object function, the noise level
implied by $v=20$ is markedly intense on purpose.

A total of 300 realizations are obtained for each of the noise levels above.
The methods to be compared are the classic cross-correlation, real-valued Jaccard
and coincidence indices, as well as the combination of cross-correlation and 
coincidence indices-based cross-correlations.  In all cases, the template signal
is a half sine identical to that shown in Figure~\ref{fig:n_0}.

\begin{figure*}[h!]  
\begin{center}
   \includegraphics[width=0.9\linewidth]{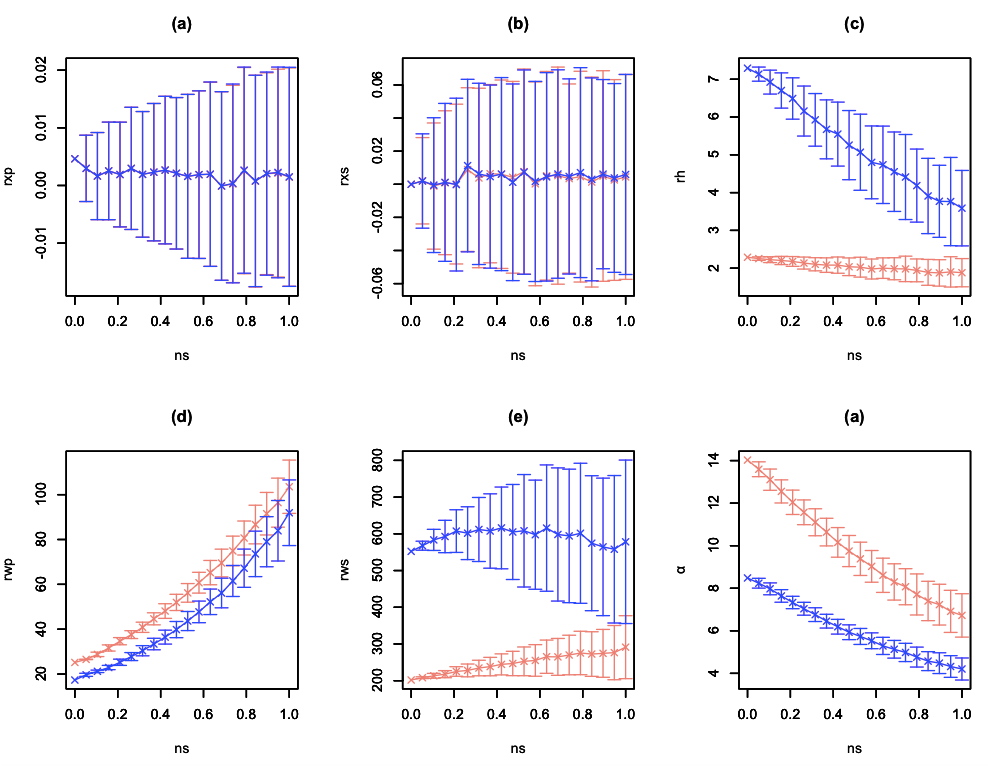}  
    \caption{The performance results in terms of the considered level noise obtained while 
    comparing the real-valued Jaccard (shown in salmon) and coincidence (in blue) methods. 
    Except for the localization of the peaks, which resulted in similar indices (b,c) in both methods, 
    the coincidence correlation approach revealed substantially superior relative performance. 
    The plots correspond to the average values for 300 realizations, $\pm$ the respective
    standard deviations.}
    \label{fig:perf_cs1}
    \end{center}
\end{figure*}

Figure~\ref{fig:perf_cs1} presents the average $\pm$ standard deviation results 
obtained respectively to the real-valued
Jaccard (results shown in salmon) and coincidence method (in blue).   Several interesting
and important results can be verified.  

First, we have that both these methods yielded quite similar performance regarding the
identification of the original position of the peaks (Fig.~\ref{fig:perf_cs1}(a,b)) 
which, in both cases,  are nearly zero, which is a quite good result.  However, when we 
move to the ratio $r_h$ shown in (c), which should correspond to a high value,
the coincidence method outperforms the real-valued Jaccard by a large margin.

The performance ratio $r_{wp}$, which should have low values, again revealed the
enhanced relative performance of the coincidence index.  The relative performance
for the index $r_{ws}$ resulted  even more striking, further corroborating the 
special characteristics of the coincidence method.

The index $\alpha$ obtained for the coincidence method, shown in Figure~\ref{fig:perf_cs1}(f), 
resulted nearly half of that observed for the real-valued Jaccard approach, again
revealing enhanced performance of the former methodology.

Though it could be expected from our preliminary analysis  that the coincidence correlation  
method would outperformed  the real-valued Jaccard correlation approach, the obtained results 
indicate that the relative advantage is surprisingly significant.  This is especially so in the case 
of the ratio $r_h$, which indicates that the coincidence method attenuates substantially the
secondary peaks.

\vspace{0.5cm}

\begin{figure}[h!]  
\begin{center}
   \includegraphics[width=1\linewidth]{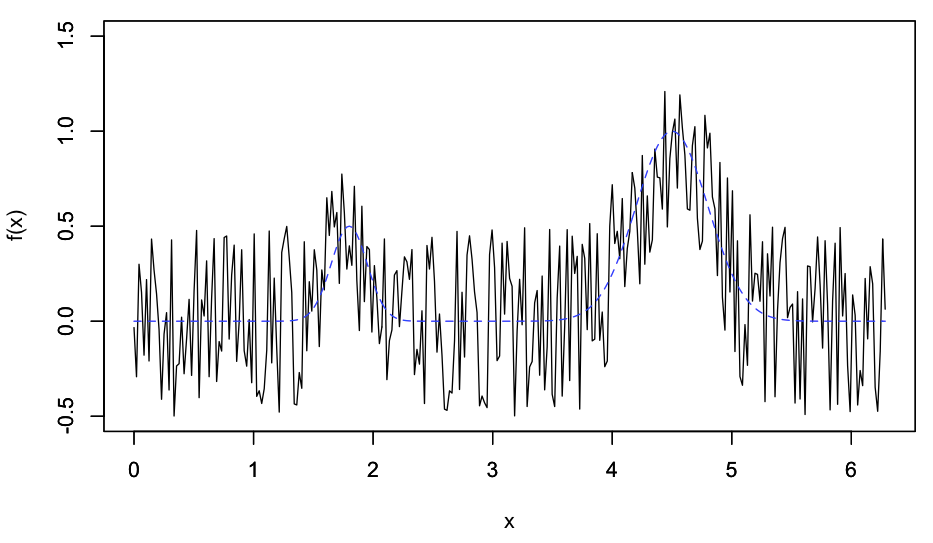}  
    \caption{Example of object function incorporating the maximum considered level
    of noise (i.e.~$=20$).  The original, noiseless version of the object function is also shown in dashed
    blue for comparison purposes.}
    \label{fig:noise}
    \end{center}
\end{figure}

Still regarding the results in Figure~\ref{fig:perf_cs1}, it is interesting to observe that 
the real-valued Jaccard and coincidence methods tend to have more similar performance
as the added noise level increases.  This is expected because too high noise levels
tend to completely undermine the patterns in the object function, as illustrated
in the object function in Figure~\ref{fig:noise} , which is respective to the highest considered
noise level.

\vspace{0.5cm}

\begin{figure*}[h!]  
\begin{center}
   \includegraphics[width=1\linewidth]{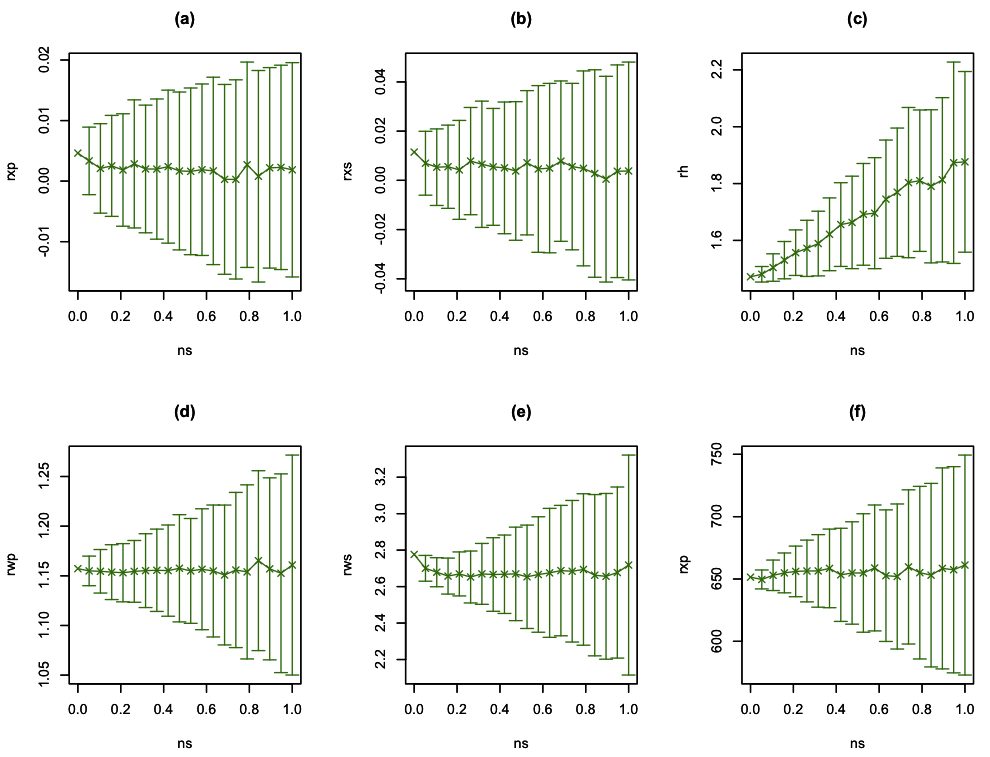}  
    \caption{Performance figures obtained for the classic correlation method. 
    Except for the localization of the peaks in (a) and (b), this method resulted almost
    hopeless regarding all other performance indices.  The plots correspond to the average 
    values for 300 realizations, $\pm$ the respective standard deviations.}
    \label{fig:perf_corr}
    \end{center}
\end{figure*}

Figure~\ref{fig:perf_corr} presents the performance figures obtained for the classic cross-correlation
method.   As already hinted by the preliminary results described in Section~\ref{sec:prelim}, 
this method is nearly
hopeless for identification of the patterns in terms of narrow, sharp peaks, while the
secondary peaks are duly attenuated.   Except for the $r_h$ index in Figure~\ref{fig:perf_corr}(c),
all other indices resulted almost constant with the noise levels.  Though a slight performance
improvement can be observed in (c), its values are far away from those allowed by the
real-valued Jaccard and coincidence methods.

\vspace{0.5cm}

\begin{figure*}[h!]  
\begin{center}
   \includegraphics[width=0.9\linewidth]{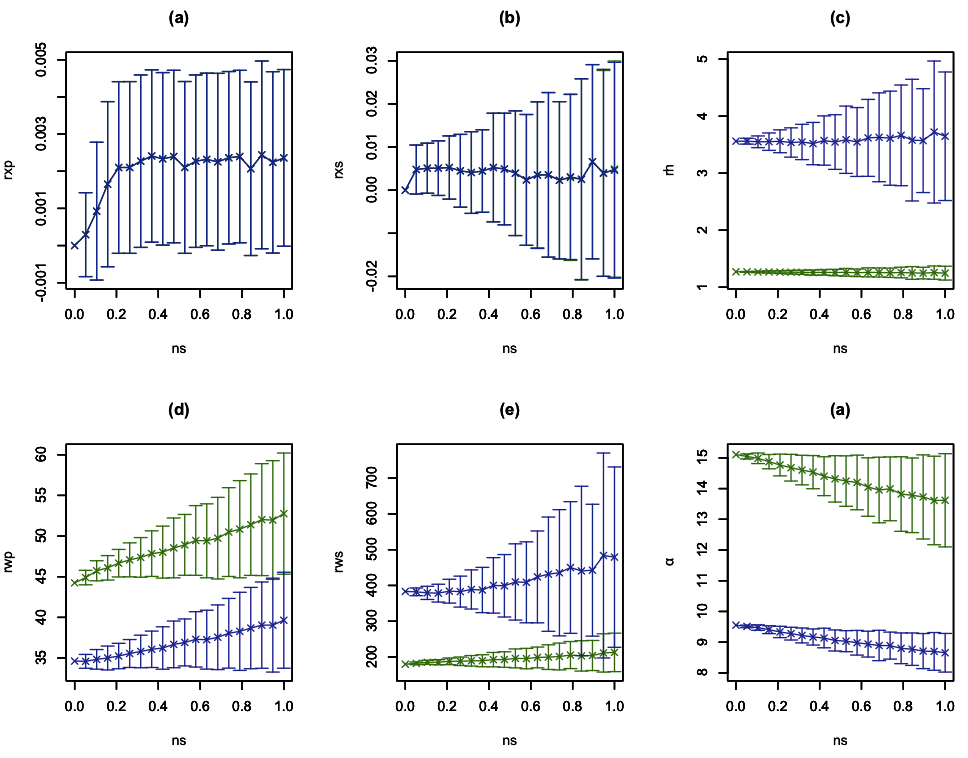}  
    \caption{Performance indices obtained for the combination of correlation prior
    to the real-valued Jaccard (dark green) and coincidence (dark blue) methods.  
    As expected, though this
    approach is less effective than the respective methods without the a priori
    correlation for low noise levels, relatively better results are observed for the largest
    levels of noise.  These results corroborate the advantage of combining correlation
    and the real-valued Jaccard or coincidence methods in the case of very noisy
    object functions.  Also of particular interest is that the several of the obtained curves
    present a markedly more linear variation with the noise level than for the other
    correlation configurations considered in this work. The plots correspond to the average 
    values for 300 realizations, $\pm$ the respective standard deviations.}
    \label{fig:perf_comb}
    \end{center}
\end{figure*}

Figure~\ref{fig:perf_comb} presents the results obtained for the combination of correlation
as preliminary processing before the application of the real-valued Jaccard or 
coincidence methods, as proposed in Section~\ref{sec:comb}.

\vspace{0.5cm}

The obtained results confirm the previous expectative that, in the average in case of object 
functions with intense level of noise it can be advantageous to apply the real-valued 
Jaccard or coincidence
correlations after the object function is filtered by the classic cross-correlation.   Indeed,
though the obtained performance figures are all worse in the case of low level noise,
they become nevertheless better than would be otherwise obtained by
not using a priori correlation in the case of the highest considered
levels of noise.  However, larger error bars are observed for more intense noise leves.

Another interesting result has been obtained in which the unfolding of the
several merit figures became much more straight or linear along the levels of noise
At the same time, the combined coincidence correlation approach resulted substantially better than
the combined real-valued Jaccard method for all levels of noise.

It should be kept in mind that these results concern situations where the template function
is smooth, being therefore capable of incorporating respective low-pass filtering action.

\begin{figure*}[h!]  
\begin{center}
   \includegraphics[width=0.32\linewidth]{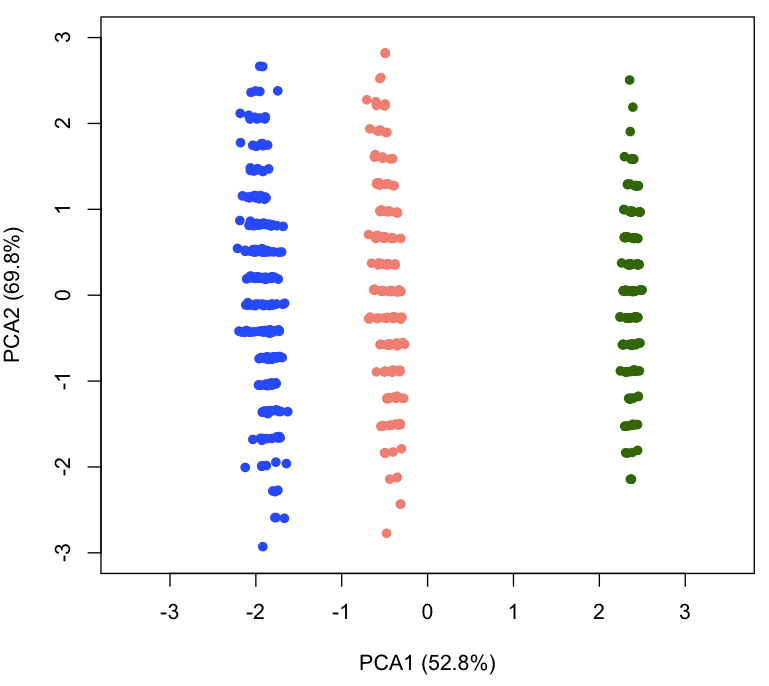}  
   \includegraphics[width=0.32\linewidth]{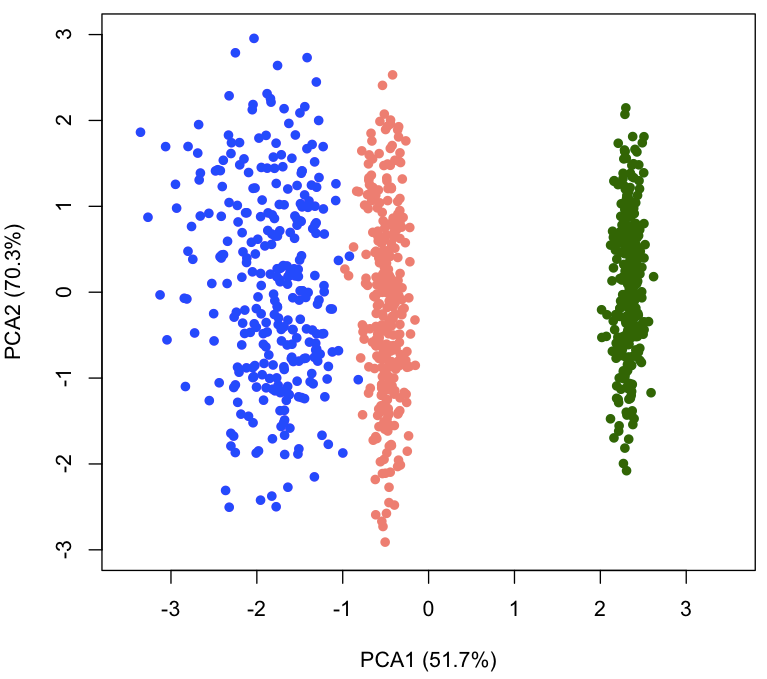}  
   \includegraphics[width=0.32\linewidth]{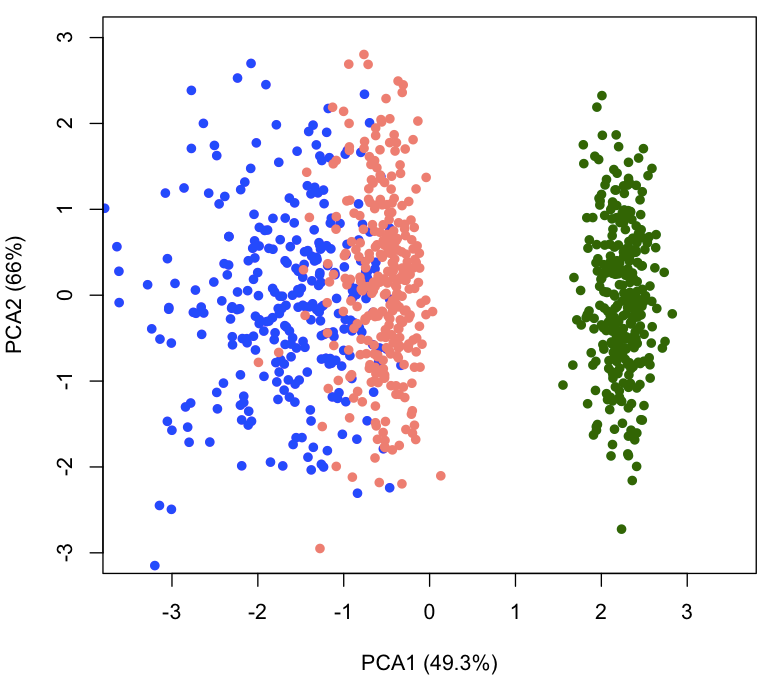}  \\
   \hspace{1cm} (a) \hspace{5.2cm} (b) \hspace{5.2cm} (c)
    \caption{Principal component analysis (PCA) of the performance figures
    obtained for the real-valued Jaccard (shown in salmon), coincidence (blue)
    and classic cross-correlation (green) for the second lowest  (a), middle (b) and highest
    considered level noises.  Observe the good variance explanation indicated in
    parentheses on the respective axes.  The fact that larger dispersions have been
    obtained for the real-valued Jaccard and coincidence indices respectively to the
    classic correlation reflects the greater ability of the two former methods,
    and especially the coincidence cross-correlation, to  capture information about the 
    original object function specificities.  These three
    PCA projections also clearly indicate the substantial relative between the 
    multiset-based correlations and the classic correlation, which is far away.  Observe
    also the confirmation that the real-valued Jaccard and coincidence methods tend
    to allow similar performance for the highest noise leves.}
    \label{fig:PCA}
    \end{center}
\end{figure*}

In order to complete our comparative performance analysis, we shown in Figure~\ref{fig:PCA}
the two main axis obtained by principal component analysis (PCA~\cite{CostaPCA,johnson:2002}) of the performance figures for the real-valued Jaccard, coincidence, and classic cross-correlation methods.  
Three PCA projections are shown, corresponding to the second lowest (a),
middle (b), and highest (c) considered noise levels.

\vspace{0.5cm}

Several interesting results are revealed from the PCA projections in Figure~\ref{fig:PCA}.
First, we have the fact that a substantial deal of the original data variance (about $70\%$ as
indicated in parentheses in the axes labels) has been explained by the two main PCA axes, which
indicates that the PCA projection is particularly representative of the original data
distribution.  Second, it becomes clear that, in all noise levels, the performance of the
classic correlation method is well separated from that of the two other multiset-based methods.

Particularly important is to observe the relative dispersions of each of the three groups,
with the coincidence index yielding the largest dispersion, followed by the real-valued 
Jaccard, and then the classic correlation approach.  These dispersions of the performance
figures reflect directly the sensitivity of each method to the specificities of the object
function, being also related to the performance of the methods, in the sense that a method
that always produce a same meaningless result independently of the shape of the object function would
have null dispersion.  Yet, relatively smaller dispersions resulted in the cases of
more typical lower noise levels considered in this work.

As expected, as the noise effectively changes the shape of the
object function, higher dispersions will be obtained by more sensitive methods
respectively to higher levels of noise.
It is also interesting to realize that it is the relative insensitivity of the classic correlation
method that paves the way to its respective combination with the real-valued Jaccard
and coincidence correlations in the case of particularly high levels of noise.

Yet, despite the larger fluctuations of the coincidence method, the respective performance
values are largely higher than those obtained by the real-valued Jaccard approach and much
less so by the extensively used classic cross-correlation.  Observe also the confirmation of
the already verified tendency of the performances of the  coincidence and real-valued Jaccard 
correlations to merge for the highest levels of noise.

\section{Concluding Remarks}

The operation of cross-correlating two functions or signals  constitutes an important mathematical  
resource that has been extensively applied for the most diverse scientific and technological
purposes.  

Though the classic cross-correlation approach relies on successive inner products between
one function displaced relatively to the other, it is also possible to apply multiset-based
approaches~\cite{CostaJaccard,CostaMset,CostaSimilarity} to derived respective
cross-correlation methodologies that have several intrinsic advantages, including being
mainly based on the non-linear binary operations (in the mathematical sense of taking
two arguments) of minimum and maximum, both of which being intuitive and requiring
low computational costs.

A set of multiset-based cross correlation approaches has been recently 
introduced~\cite{CostaJaccard,CostaMset,CostaSimilarity} that includes the real-valued
Jaccard, interiority, coincidence, addition-based real-valued Jaccard, as well as the
addition-based coincidence correlations.  Preliminary results~\cite{CostaMset,CostaSimilarity}
indicated that these operations, except the interiority index which is used as an ancillary
resource in the definition of the coincidence indices have, when compared with the
classic cross-correlation approach, substantially enhanced potential for
selective and detailed similarity quantification, including in tasks such as filtering and
template matching.   

In the present work, we set out at developing a comprehensive and systematic comparison
approach between the performance allowed by the above mentioned cross-correlation
methods with respect to the important pattern recognition task of template matching, which
can be readily understood as a kind of filtering.

After presenting the adopted multiset concepts and respectively derived cross-correlation
approaches, we performed a preliminary comparison aimed at providing an overall
appreciation of the relative potential of the several considered methods.  The results
clearly corroborated the enhanced potential of the multiset methods respectively to the
classic cross-correlation.  In addition, it was verified that the addition-based methods
led to similar performances to those obtained by the real-valued Jaccard approach.  
These preliminary results also indicated that the interiority approach is
not particularly powerful when considered without being combined with the Jaccard
index in order to obtain the coincidence method.   These findings allowed us to
narrow the focus of our comparison on three main alternative cross-correlation 
methods: the real-valued Jaccard, coincidence, and classic cross-correlation approaches.

In addition, the identification of the enhanced robustness of the classic cross-correlation
to intense noise levels observe in the preliminary comparison also motivated the proposal
of a method in which the cross-correlation is performed prior to the application of the
multiset-based mehtods as means to reduce the noise as a preparation for the
matching detection.

In order to characterize and understand better the respective performance of the
three chosen methods, we proposed a formal and comprehensive framework to be
used for systematic performance comparison.  This framework, which can also be understood
as an additional contribution of the present work, takes into account several important
quantitative indices expressing the performance of the cross-correlation with respect to
several relevant features characterizing good performance.

The application of the proposed comparison framework led to several interesting
results.

First, we have that the real-valued Jaccard and coincidence methods allowed the best
overall template matching performance respectively to the classic cross-correlation, 
therefore corroborating further the preliminary expectations.  The performance was also
observed not to be significantly undermined even by intense levels of additive noise.

Among the two multiset-based approaches, the coincidence method resulted significantly 
more effective, leading to best performance with respect to virtually every considered index.  
The proposed cross-correlation method combining the classic cross-correlation before
the application of the multiset-based methods was also found to constitute an interesting
alternative in cases of intense noise.

Additional characterization of the obtained performance indices by using principal
component analysis provided further understanding the potential of each investigated
method.  It was therefore confirmed that the real-valued Jaccard and coincidence 
cross-correlation methods have performance markedly distinct from that of the
classic cross-correlation.  In addition, the sensitivity of the two multiset-based
methods was confirmed by the observation of larger respective dispersions in the
PCA space that are consequence of their enhanced sensitivity to the modifications
implied by the noise on the specific shape of the object function.  These complementary
results also substantiated that the coincidence method provided the best overall
performance at least for the type of data and problem addressed in our comparative
analysis.

The reported concepts, methods, and results have important and wide implications for
many areas dependent of the cross-correlation or the related convolution operations,
including signal processing (e.g.~\cite{oppenheim:2009,Lathi,Proakis}), 
pattern recognition (e.g.~\cite{DudaHart,Koutrombas,brigham:1988}), shape analysis
(e.g.~\cite{shapebook}), complex networks (e.g.~\cite{CostaCComplex,surv_appl,networks:2010})) 
as well as deep learning (e.g.~\cite{Schmidhuber,Hinton}), among many other fields.

One point of particular interest is to contemplate to which 
extent similar methods could be employed in natural recognition systems involving 
neuronal cells, as motivated by their relative simplicity and low computational cost.  
An interesting related issue regarding the possible eventual neuronal operations
that would be correlate to the multiset-based methods addressed in the present work.

Given the importance of similarity concepts and methods in the physical sciences, not
to mention virtually every other scientific and technological fields, the implications of the
methods results reported in the present work are particularly ample, paving the way
to a large number of further developments.  Indeed, every current concept or method
based on similarity, inner products, and/or convolution and correlation can be re-assessed
in the light of the presented results indicating the enhanced performance of the Jaccard and
multiset-derived methods relatively to the extensively applied classic cross-correlation,
as well as related concepts including the cosine similarity and inner product.

As more immediate future developments more close and specifically related to the issues addressed in the
present work,  it would be interesting to consider other types of functions and cross-correlation based methods, higher dimensional vector and function spaces, as well as to study the effect of the relative 
function magnitudes and other types of noise 
on the results.  Related research is being conducted and results are to be published opportunely.

\vspace{0.7cm}
\emph{Acknowledgments.}

Luciano da F. Costa
thanks CNPq (grant no.~307085/2018-0) and FAPESP (grant 15/22308-2).  
\vspace{1cm}

\bibliography{mybib}
\bibliographystyle{unsrt}

\end{document}